\documentclass{article}
\usepackage{spconf,amsmath,graphicx}
\usepackage{hyperref} 
\usepackage{booktabs}
\usepackage{amssymb,stackengine,graphicx}   
\usepackage{multirow}
\usepackage{pifont}
\usepackage{adjustbox}
\usepackage{mathtools, cuted}
\usepackage{xcolor}
\hypersetup{
    colorlinks,
    linkcolor={black},
    citecolor={black},
    urlcolor={black}
}
\definecolor{mygray}{gray}{0.6}

\newcommand{\cmark}{\ding{51}}%
\newcommand{\xmark}{\ding{55}}%

\title{Paralinguistics-Enhanced Large Language Modeling of \\ Spoken Dialogue}
\name{
\begin{tabular}{@{}c@{}}
Guan-Ting Lin$^{\dagger 1,2}$\thanks{$^{\dagger}$Work done during first author's internship with Amazon Alexa AI.}\quad Prashanth Gurunath Shivakumar$^{1}$\quad Ankur Gandhe$^{1}$\quad Chao-Han Huck Yang$^{1}$ \\ Yile Gu$^{1}$ \quad Shalini Ghosh$^{1}$\quad Andreas Stolcke$^{1}$\quad Hung-yi Lee$^{2}$\quad Ivan Bulyko$^{1}$
\end{tabular}
}

\address{$^{1}$Amazon Alexa AI, USA\\ $^{2}$National Taiwan University, Taiwan}
\begin{document}
\ninept
\maketitle
\begin{abstract}
Large Language Models (LLMs) have demonstrated superior abilities in tasks such as chatting, reasoning, and question-answering. However, standard LLMs may ignore crucial paralinguistic information, such as sentiment, emotion, and speaking style, which are essential for achieving natural, human-like spoken conversation, especially when such information is conveyed by acoustic cues. We therefore propose \textbf{Paralin}guistics-enhanced \textbf{G}enerative \textbf{P}retrained \textbf{T}ransformer (\textbf{ParalinGPT}), an LLM that utilizes text and speech modalities to better model the linguistic content and paralinguistic attributes of spoken dialogue. The model takes the conversational context of text, speech embeddings, and paralinguistic attributes as input prompts within a serialized multitasking multimodal framework. Specifically, our framework serializes tasks in the order of current paralinguistic attribute prediction, response paralinguistic attribute prediction, and response text generation with autoregressive conditioning. 
We utilize the Switchboard-1 corpus, including its sentiment labels as the paralinguistic attribute, as our spoken dialogue dataset. 
Experimental results indicate the proposed serialized multitasking method outperforms typical sequence classification techniques on current and response sentiment classification. Furthermore, leveraging conversational context and speech embeddings significantly improves both response text generation and sentiment prediction. Our proposed framework achieves relative improvements of 6.7\%, 12.0\%, and 3.5\% in current sentiment accuracy, response sentiment accuracy, and response text BLEU score, respectively.


\end{abstract}

\section{Introduction}
\label{sec:intro}
Large language models (LLMs)~\cite{openai2023gpt4} have enabled substantial recent progress in dialogue generation, language understanding, and reasoning, mostly operating on text. However, the most natural form of communication for human-human interaction is speech. Speech signals contain a wealth of information related to human communication, encompassing linguistic aspects (such as words, phonetics, syntax, and semantics), paralinguistic elements (including emotions, sentiments, and speaker characteristics), and pragmatic factors (like sarcasm and attitude).

\begin{figure}[t]
  \centering
  \includegraphics[width=0.85\linewidth]{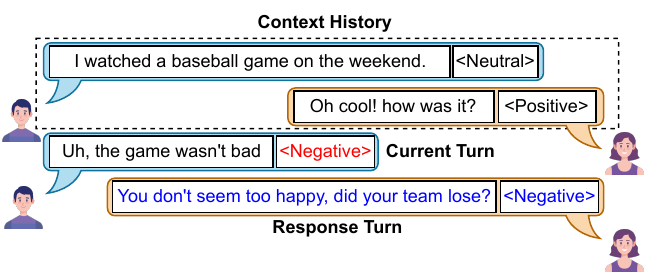}
  \vspace{-0.4cm}
  \caption{Speech dialogue scenario. \texttt{<>} denotes paralinguistic sentiment associated with an utterance. ``Context History" refers to conversational turns before the ``Current Turn", and ``Response Turn" is what follows the current turn. }
  \vspace{-0.4cm}
  \label{fig:scenario}
\end{figure}

In human-machine interaction environments, deployment of LLMs that solely rely on lexical aspects of speech thus fail to capture vital information present in human spoken dialogue.
Consequently, LLMs struggle to perceive and differentiate between speaking styles, often responding to users in a monotonous manner~\cite{zhao2023chatgpt}. Furthermore, there are instances where subtle differences in speaking styles can alter the semantic meaning of an utterance, such as with sarcasm~\cite{castro2019towards}. Consider the example shown in Figure~\ref{fig:scenario}: when the current turn is "\textit{Umm ... I think it wasn't a bad game}" conveyed with a \textit{negative} tone, the textual information still leans towards a positive or neutral sentiment. By ignoring the speech signal the actual sentiment of the speaker would be misintepreted.

Despite prior work that leverages sentiment or emotion labels to enhance dialogue modeling~\cite{bothe2017dialogue, dias2022towards, song-etal-2019-generating, firdaus2020emosen, varshney2021modelling, liu-etal-2021-modulating, zhao2023chatgpt}, such studies are primarily based on textual dialogue data, with sentiment or emotion labels being inferred from the word alone. On the other hand, prior work in modeling spoken dialogue has also shown that prosodic information is helpful for language modeling, i.e., predicting the words~\cite{stolcke2000dialogue, WARD2012161}.  

In this work, moving beyond information extracted from text alone, we introduce \textbf{ParalinGPT}, a language model capable of incorporating speech embeddings from a self-supervised speech encoder as input prompts, within a \textit{serialized multitasking} multimodal framework designed to \textbf{predict both upcoming text- and speech-based paralinguistic information}. We employ the Switchboard-1 dataset, the largest collection of human-human spontaneous spoken dialogues with annotated ``speech sentiments", as our primary spoken dialogue data source.

\begin{figure}[t]
  \centering
  \includegraphics[width=0.95\linewidth]{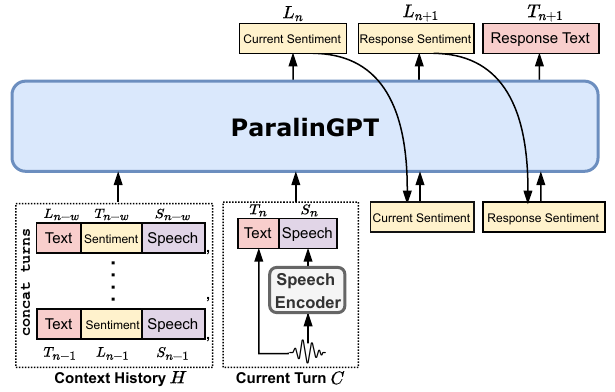}
  \vspace{-0.2cm}
  \caption{\texttt{ParalinGPT} and serialized multitasking: The history context (including text, speech, and sentiment labels), current text, and current speech encoding are the input prompt for the ParalinGPT LLM. The prediction target is the current sentiment label, the response sentiment label, and the response text via autoregression.}
  \vspace{-0.4cm}
  \label{fig:prosodyGPT}
\end{figure}

To our knowledge, this is the first attempt to employ a multimodal language model for joint response text generation and prediction of response and current sentiment labels, particularly in the domain of real-world spoken dialogues. Our specific contributions include:
\begin{itemize}
    \item Use of LLM to infer current sentiment, predict response sentiment and use both to predict the textual response, using an unified serialized multitasking framework.
    \item Demonstration that utilizing speech features and context history sentiment labels improves current and response sentiment prediction, as well as response text generation, in natural human-human speech dialogue.
    \item Results showing that the framework achieves superior performance compared to sequence classification baselines, with 6.7\% and 12.0\% accuracy improvement for current and response sentiment, respectively, while seeing 3.5\% relative improvement in BLEU score for response text generation. 
\end{itemize}

\section{Related Work}

\subsection{Dialogue Modeling with Sentiment and Emotion}
There are several prior studies aiming to model sentiment~\cite{bothe2017dialogue, dias2022towards, li2018emotion, ghazarian2021user, ghazarian2022wrong, firdaus2022polise} and emotion~\cite{li2018emotion, song-etal-2019-generating, firdaus2020emosen, varshney2021modelling, liu-etal-2021-modulating} in dialogue. Most of this work concentrates on current-turn sentiment/emotion as a pure classification task. However, previous work has also shown that using multimodal approaches is crucial for sentiment and emotion analysis~\cite{li2019acoustic, motamed2017speech, yan2016multi, poria2017ensemble, lin2023utility, lakomkin2019incorporating}.

There have been fewer attempts at response modeling where the goal is to predict the response sentiment~\cite{bothe2017dialogue} or emotion~\cite{li2018emotion, shi20h_interspeech, shi23e_interspeech} label; others are leveraging the sentiment~\cite{firdaus2022polise, dias2022towards} or emotion label~\cite{song-etal-2019-generating, firdaus2020emosen, varshney2021modelling, liu-etal-2021-modulating, zhao2023chatgpt} to guide the response text generation. It is worth noting that Varshney \textit{et al.}~\cite{varshney2021modelling} proposed a multitask framework with context-turn emotion recognition and response generation, showing improved emotional relevancy of the generated response, but they do not predict the response emotion. Compared to this prior work, our proposal is novel in the following aspects: (i) we use multimodal input (speech and text) with multi-turn dialog context on real-world spoken dialogue, and (ii) we model current and response sentiment, as well as response text as conditioned on the paralinguistic labels.

\subsection{Language Modeling for Speech}
Much recent research uses spoken language modeling on discrete speech/audio units, either using a decoder-only LM to perform speech continuation~\cite{lakhotia2021generative, kharitonov2022text, nguyen2023generative, borsos2023audiolm} 
or leveraging paired transcripts and speech for ASR and TTS tasks~\cite{rubenstein2023audiopalm, zhang2023speechgpt}. However, discrete audio/speech units mainly capture content information and lack prosodic and paralinguistic information. pGSLM~\cite{kharitonov2022text} explicitly models extra pitch and duration units to improve speech naturalness. Notably, dGSLM~\cite{nguyen2023generative} is the only study of two-channel language modeling for spoken dialogue, but the generated speech is still lacking in semantic coherence. Different from the above-mentioned studies, we focus on equipping a strong text LM with \textit{continuous} speech embeddings to capture the paralinguistic signal in spoken dialogue.

\section{Method}

\subsection{The Serialized Multitasking Framework}
In the course of spontaneous natural human conversation, when the next speaker prepares to respond, a natural progression involves understanding the ongoing turn (including current sentiment), selecting an appropriate response style (response sentiment prediction), and subsequently crafting the response content (response text generation).

Inspired by the above rationale, and to move beyond isolated predictions, we propose a sequential conditioning mechanism, encompassing current sentiment prediction, response sentiment prediction, and response text generation, all joined in  an autoregressive  chain; we call this framework \textbf{serialized multitasking} since the model's input and output streams encode multiple tasks. The proposed framework is illustrated in Figure \ref{fig:prosodyGPT}.

We formalize the problem as follows: given a multi-turn spoken dialog with historical context turns $H$ and window size $w$, comprising speech $\{S_{n-w},.., S_{n-1}\}$, text $\{T_{n-w},.., T_{n-1}\}$, and sentiment labels $\{L_{n-w},.., L_{n-1}\}$, a current turn $C$ comprising speech $S_{n}$ and text $T_{n}$, our prediction target $O$ is the current sentiment $L_{n}$, response sentiment $L_{n+1}$, and response text $T_{n+1}$. The overall objective is as follows:
\begin{eqnarray}
    P(O|H, C) & = & P(L_{n}|H, C) P(L_{n+1}|H, C, L_{n}) \nonumber \\
            & &   \quad P(T_{n+1}|H, C, L_{n}, L_{n+1}), 
\end{eqnarray}
where $H = \{F_t(T_{n-w}), F_t(L_{n-w}), F_s(S_{n-w}), ..., F_t(T_{n-1}), \\F_t(L_{n-1}), F_s(S_{n-1})\}$, and $C = \{F_t(T_{n}), F_s(S_{n})\}$. $F_s$ is a utterance-wise speech encoder, $F_t$ is the subword embedding. The sentiment labels, $L$, are incorporated into the text using special markers \texttt{<}$L$\texttt{>}. We treat the concatenation of $(H, C)$ as the input prompt, and directly feed the continuous embedding into the LM input space.

The training objective is the typical cross-entropy loss for language modeling with teacher forcing. During testing, we generate $O$ by autoregressively feeding earlier predicted tokens back into the LLM inputs to predict a target sequence that contains current sentiment, response sentiment, and response text all represented as text. Note that the text and sentiment label in the history $H$, as well as the current text tokens $T_n$ are ground truth. Future work can look at using ASR output and erroneous versions of historical sentiment labels. After sequence generation, we use the special marker \texttt{<}\texttt{>} to decode sentiment predictions for evaluation. 


\subsection{Language Model}
Due to the recent success of autoregressive LLMs, we focus on the decoder-only transformer model for our purposes. We utilize DialoGPT~\cite{zhang2020dialogpt} as our pretrained LM, which is a GPT2 model fine-tuned on large amounts of text dialogue (Reddit posts), with human-level dialogue generation capability.

\begin{table*}[t]
\centering
\caption{Experimental results: ``Curr" and ``Res" denote current and response sentiment, respectively. In the ``Modality" column, ``T'' indicates text, and ``S'' denotes speech. ``\cmark" indicates presence and ``\xmark" indicates absence of context information. As sequence classification baselines ($^\ast$) we train separate models for current sentiment (Curr UA) and response sentiment (Resp UA).}
\adjustbox{width=0.85\textwidth}{
\begin{tabular}{c|c|c|cc|ccc}
\toprule
\multirow{2}{*}{\textbf{Method}} & \multirow{2}{*}{\textbf{\#}}  & \multicolumn{1}{c|}{\textbf{Current}}   & \multicolumn{2}{c|}{\textbf{Context History}} & \multicolumn{3}{c}{\textbf{Prediction Target}}            \\ \cline{3-8} 
                          & & \textbf{Modality} & \textbf{Modality} & \textbf{Sentiment} & \textbf{Curr UA ($\uparrow$)} & \textbf{Resp UA ($\uparrow$)}  & \textbf{BLEU ($\uparrow$) }         \\ \midrule
                        
Lu et al.~\cite{lu2020speech}         & 1     & E2E ASR     & \xmark           & \xmark             &    62.4                    & - & -\\
Shon et al.~\cite{shon21_interspeech} & 2       & T           & \xmark          & \xmark             &    64.5                   & -& - \\
Human performance~\cite{lu2020speech}    & 3          & \xmark     & \xmark           & \xmark             &    84.6          & -          &  - \\ \midrule
Random baseline                 & 4    & \xmark    & \xmark       & \xmark             & 33.3             & 33.3            & -\\ \midrule

\multirow{2}{*}{LM only}    & 5 & T   & \xmark       & \xmark            & -         & -      & 14.2      \\
                          
                          & 6   & T & T           & \xmark             & -  & -             & 14.5        \\ \midrule
\multirow{3}{*}{Sequence Classification Baseline$^*$}  & 7 & T                   &  \xmark     &  \xmark      &   66.3          &   36.1        & -\\ 
 & 8 & S          &  \xmark     &  \xmark      &   65.1           &   34.8        & -\\
  & 9  & T + S              &  \xmark     &  \xmark      &   68.1           &   37.4      & -\\ \midrule

\multirow{6}{*}{Proposed Serialized Multitasking Framework}    & 10   & T  & \xmark           & \xmark            & 68.3           & 38.0           & 14.0          \\
                        & 11     & T & T          & \xmark             & 69.3           & 40.5             & 14.4\\
                        & 12     & T & T           & \cmark             & 70.3  & 41.4   & 14.5 \\
& 13  & T + S & \xmark           & \xmark            & 70.7           & 38.6           & 14.1      \\
                  & 14          & T + S  & T + S           & \xmark             & 71.3   & 40.9            & 14.7 \\ 
                  & 15           & T + S  & T + S           & \cmark           & \textbf{71.7}          & \textbf{41.9}    & \textbf{14.9}\\
\bottomrule
\end{tabular}
}
\vspace{-0.4cm}
\label{tab:all}
\end{table*}

\subsection{Speech Encoder}
We use the wav2vec 2.0-large~\cite{baevski2020wav2vec} self-supervised model pretrained on Switchboard, Fisher, LibriLight~\cite{hsu2021robust} as the speech encoder. Wav2vec 2.0 has been shown to encode rich prosodic information useful for speech sentiment and sarcasm detection~\cite{lin2023utility}. We use the middle layer (the 12th of 24 layers) of the wav2vec 2.0 model for frame-wise feature extraction, since the middle layer achieved best results in our initial linear probing experiments with sentiment analysis, aligned with the finding in \cite{li2023exploration}. We apply mean-pooling and a linear feature projector (1.04M parameters, 0.3\% of all LM parameters) to extract utterance-wise embeddings. To reduce computational cost, we freeze the parameters of wav2vec 2.0, and only update the linear projector during training. 


\vspace{-0.2cm}
\section{Experiments}
\vspace{-0.2cm}

\begin{table}[tb]
\centering
\caption{Dataset statistics. The ``Class" row lists the distribution of positive/negative/neutral classes. }
\adjustbox{width=0.49\textwidth}{
\begin{tabular}{c|ccc}
\toprule
\textbf{}                  & \textbf{Train}                       & \textbf{Dev} & \textbf{Test} \\ \midrule
\# samples                 & 43600                                & 1158         & 1319          \\
\# dialogues               & 1124                                 & 56           & 56            \\
\# speakers                & 385                                  & 86           & 72            \\ 
\multicolumn{1}{c|}{Class} & \multicolumn{1}{c}{13,074/7,317/23,209} & 349/222/587  & 316/190/813  \\\bottomrule
\end{tabular}
}
\vspace{-0.3cm}
\label{tab:data}
\end{table}

\subsection{Data and Metrics}
We use the Switchboard-1 corpus~\cite{godfrey1992switchboard} (LDC97S62), along with speech sentiment annotations  (LDC2020T14)~\cite{chen2020large}. The sentiment labels were annotated through a three-way human agreement process, categorizing utterances into positive, neutral or negative. This dataset encompasses a total of 140 hours of audio. By utilizing the temporal information for the starts and ends of audio segments, we are able to retrieve the context history of each conversational turn. Since the data split from \cite{chen2020large} is not released, we randomly sample 5\% of conversations without speaker overlap as holdout validation and testing sets. It is important to emphasize that our dataset division adheres to a \textit{strict dialogue-level separation without speaker overlap}. The details of data statistics are listed in Table~\ref{tab:data}. We normalize the text data to lowercase and remove nonlexical markers such as laughter and noises. 


For evaluating sentiment classification, similar to \cite{chen2020large}, we use unweighted accuracy (UA) for both current and response sentiment prediction, to account for the sentiment class imbalance in the dataset.
To evaluate response text generation, we use the BLEU (4-gram) score to measure the content similarity of the generated text response to the ground truth. 

\vspace{-0.5cm}
\subsection{Implementation Details}
\vspace{-0.1cm}
We utilize the pretrained DialoGPT with 345M parameters \cite{DialoGPT}
as LM and wav2vec 2.0 Large as the speech encoder \cite{Wav2vec2.0}.
For the context history, we use window size $w = 4$ and truncate the input if the length exceeds 320 tokens during training. We use a batch size of 32 and Adam optimizer with learning rate 0.0001, and train for 100k steps. During training, the parameters of the speech encoder are frozen and only the parameters of the LM and speech linear projector are updated. We select the best model according to the validation set, and report the performance on the test set. 
\vspace{-0.3cm}
\subsection{Baseline Approaches}
For each task, we consider several baselines against which to benchmark our proposed model, based on published results.\\
\textbf{Response text generation:} We fine-tune DialoGPT on Switchboard data using the standard LM objective and multi-turn context history as a baseline.
\\
\textbf{Sentiment classification:} We use three baseline systems to benchmark sentiment classification against our proposed technique. (i) \textit{Random baseline}, since ours is the first work targeting response sentiment prediction on the Switchboard corpus. (ii) \textit{Sequence classification}: for speech, we fine-tune wav2vec-2.0 with a classifier head; for text, we fine-tune DialoGPT for simple classification; for multimodal approaches combining text and speech we augment DialoGPT with input text concatenated with wav2vec 2.0 mean-pooled embeddings. (iii) \textit{Prior published results} on the Switchboard corpus for utterance classification.

\section{Results}


\subsection{Baselines} 
From prior work on single-utterance sentiment analysis, we quote the performance reported in \cite{lu2020speech, shon21_interspeech}, while noting differences in terms of the data split. For response sentiment, we utilize a random (chance) baseline, since there is no prior work on Switchboard to compare to. The sequence classification baseline is used to assess the task difficulty and for comparison to our proposed serialized multitasking method.

From the results in Table~\ref{tab:all}, the Text+Speech (row 9) approach consistently yields superior performance compared to the Text-only (row 7) and Speech-only (row 8) approaches. This suggests that for both current and response sentiment prediction, speech and text modalities complement each other.
For response sentiment prediction, the performance stands at 37.0\% UA, which is considerably lower than the performance for current sentiment classification, as the latter represents a more challenging task (since the corresponding words and speech are not known yet). Overall, our sentiment classification baseline reaches comparable and even better performance than the prior studies, and yields nontrivial response sentiment prediction compared to the random baseline.    

For evaluations of response text generation, we use a domain-adapted DialoGPT model as baseline (rows 5 and 6 in Table~\ref{tab:all}). It achieves a BLEU score of 14.2 and 14.5 without and with multi-turn context, respectively.

\subsection{Serialized multitasking framework}
Comparing the proposed method (Table~\ref{tab:all}, row 10) to the classification baseline (row 7), we observe better performance for both current and response sentiment, with 3.0\% relative UA improvement for current sentiment and 2.4\% relative improvement for response sentiment. These results suggest that joint sentiment and language modeling, in the correct order, can lead to improvements beyond relying on sequence classification alone.
However, for response text generation, we see a slightly worse BLEU score compared to the LM-only baseline (rows 10 and 5). One possible reason for this is that the response sentiment prediction is noisy which can affect the subsequent text generation.

We also experiment with varying the ordering of tasks in the serialized multitasking framework, as shown in Table~\ref{tab:ablation}. We find that predicting the current sentiment followed by response sentiment and finally the response word tokens yields the best results. This suggests that response sentiment can be useful to guide the generation of response text. The response sentiment prediction also benefits by conditioning on the current-utterance sentiment prediction. Besides, when current sentiment prediction is absent, we see a slight performance drop compared to the proposed method, which shows that understanding current sentiment is helpful to generate proper responses. 

\subsection{Multimodal serialized multitasking framework}
Additionally, incorporating speech embeddings improves both current and response sentiment classification, with a 3.5\% and 1.6\% relative UA improvement, respectively. These results suggest that the inclusion of the speech modality provides valuable nonlexical information, such as laughter and expressive speaking styles, which complement the text modality effectively.

\begin{table}[t]
\centering
\caption{Results breakdown with respect to task ordering for the text modality. $L_n$: current sentiment label, $L_{n+1}$ is the response sentiment label, and $T_{n+1}$ is the response text.
}
\label{tab:ablation}
\adjustbox{width=0.42\textwidth}{
\begin{tabular}{c|cll}
\toprule
\textbf{Task Order}        & \textbf{Curr UA} & \textbf{Resp UA}& \textbf{BLEU} \\ \midrule
$L_n \rightarrow L_{n+1} \rightarrow T_{n+1} $ (Proposed)                            & 70.3             & 41.4                    & 14.5          \\
$L_n \rightarrow T_{n+1} \rightarrow L_{n+1}$ &   64.6     &        38.8               &    13.1           \\
$L_{n+1} \rightarrow T_{n+1} $         &    -             &         40.9           &    14.4           \\
\bottomrule   
\end{tabular}
}
\end{table}

\begin{figure}[t]
\centering
\includegraphics[width=0.65\linewidth]{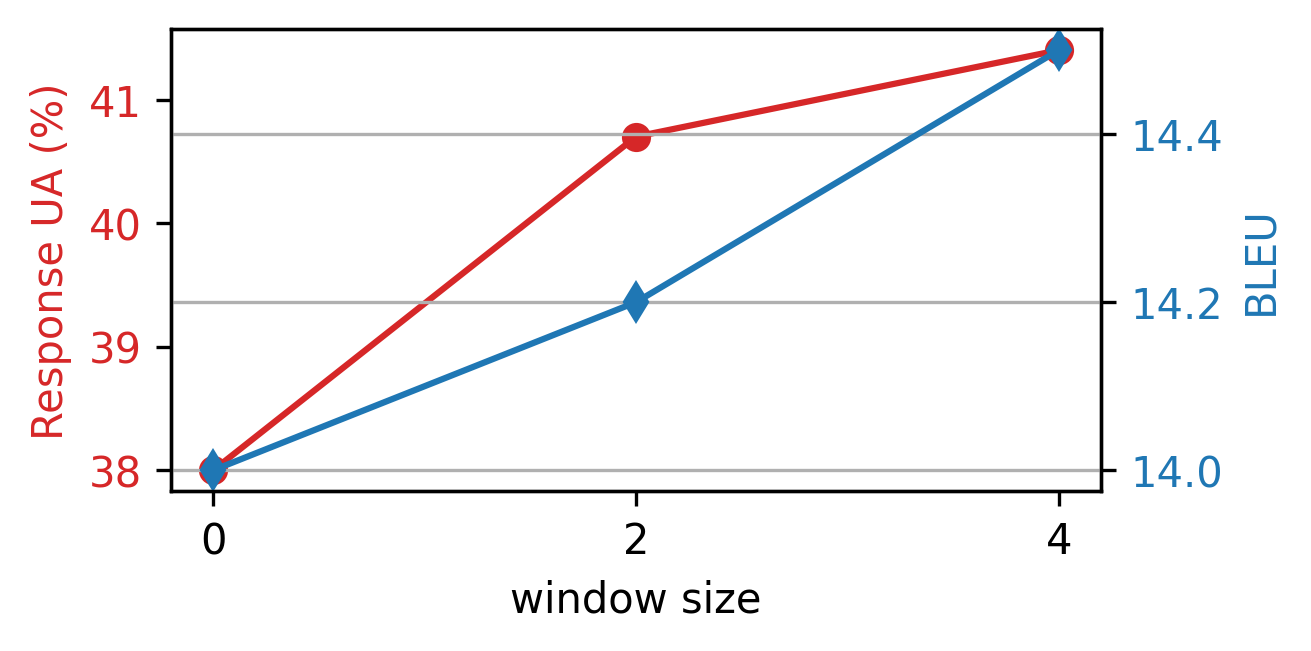}
  \vspace{-0.3cm}
  \caption{Effect of context length on response sentiment classification and text generation with the proposed method.}
  \vspace{-0.5cm}
  \label{fig:window}
\end{figure}

\subsection{Role of Multi-turn Context}
Further, we leverage multi-turn context history in the form of text, speech and sentiment information (see column "Context History" in Table~\ref{tab:all}).\\
\textbf{Textual context}: Text context is crucial for both the LM-only baseline and the proposed approach. We observe a 2.1\% relative BLEU score improvement in the LM-only baseline. Moreover, for the text serialized multitasking method, textual context also contributes to improved sentiment prediction. Besides a 2.9\% relative BLEU score gain, both current and response sentiment prediction is improved by 1.5\% and 3.4\% relative UA, respectively. As for the \textit{context history length}, from Figure~\ref{fig:window}, we tried smaller context window sizes ($w=2,0$) and observed that performance drops gradually across all prediction targets, highlighting the importance of longer contexts.\\
\textbf{Multimodal context}: Beyond text context, we can incorporate speech embeddings for contextual turns to provide more acoustic clues for interpretation and prediction. Comparing text (Table~\ref{tab:all}, row 11) with Text+Speech context (row 14), we observe further improvements of 1.4\%, 0.5\%, and 2.1\% in relative terms for current sentiment, response sentiment, and BLEU score, respectively.\\
\textbf{Sentiment context}: Lastly, using contextual sentiment labels also enhances the overall performance of the proposed method. For the text modality, there are relative gains of 1.4\%, 1.7\%, and 0.7\% in current sentiment, response sentiment, and response text, respectively. Overall, the best-performing multimodal serialized method (row 15) achieves a current sentiment accuracy of 71.7\%, response sentiment accuracy of 41.9\%, and a response text BLEU score of 14.9. This result implies that contextual speech embeddings and sentiment information are complementary. We hypothesize that the rich information in speech, such as timing and intonation, helps to  disambiguate and understand the spoken dialogue more effectively.

\section{Conclusion}
We have introduced \textbf{ParalinGPT}, a modeling framework for spoken dialogues that integrates paralinguistic and linguistic information in a single autoregressive model. This framework enables the joint generation of response text and current and response sentiment prediction. Through our experiments on a corpus of human-human conversations, we show the effectiveness of the proposed serialized multitasking method compared to separate classification and language models. Additionally, we demonstrate that incorporating self-supervised speech features and context history further improves performance. 
There are several directions for future research, including optimizing the time resolution and pooling method for paralinguistic embeddings, using the latest large-scale LLMs, and adding more tasks under serialized multitasking, such as turn taking prediction and emotion recognition. 




\clearpage

\bibliographystyle{IEEEbib}
\footnotesize
\bibliography{refs}

\end{document}